\documentclass[10pt,twocolumn,letterpaper]{article}

\usepackage{cvpr}

\usepackage{xspace}
\usepackage{multicol}
\usepackage{multirow}
\usepackage{marvosym}
\usepackage{pifont}
\usepackage[table,dvipsnames]{xcolor}

\makeatletter
\DeclareRobustCommand\onedot{\futurelet\@let@token\@onedot}
\def\@onedot{\ifx\@let@token.\else.\null\fi\xspace}

\makeatother

\definecolor{gen_blue}{RGB}{99,113,250}
\definecolor{gen_red}{RGB}{239,99,75}
\definecolor{gen_green}{RGB}{0,180,139}
\definecolor{gen_gray}{RGB}{165,165,165}
\definecolor{cvprblue}{rgb}{0.21,0.49,0.74}

\usepackage[pagebackref=false,breaklinks,colorlinks,allcolors=cvprblue]{hyperref}

\usepackage[accsupp]{axessibility}

\def\paperID{}
\def\confName{CVPR}
\def\confYear{2026}

\title{\vspace{-0.7cm}Not All Points Are Equal: Uncertainty-Aware 4D LiDAR Scene Synthesis}

\author{
Xiang Xu$^{1}$ \quad Alan Liang$^{2}$ \quad Youquan Liu$^{3}$ \quad Xian Sun$^{4}$ \quad Linfeng Li$^{2}$ \quad Lingdong Kong$^{2}$
\\
Ziwei Liu$^{5}$ \quad Qingshan Liu$^{6,7}$
\\[0.5ex]
{\small$^1$NUAA \quad $^2$NUS \quad $^3$FDU \quad $^4$Duke \quad}
{\small$^5$NTU \quad $^6$NJUPT \quad $^7$SKL-TI}
}

\begin{document}

\twocolumn[{
    \renewcommand\twocolumn[1][]{#1}
    \maketitle
    \begin{center}
    \centering
    \captionsetup{type=figure}
    \vspace{-0.5cm}
    \includegraphics[width=0.9\linewidth]{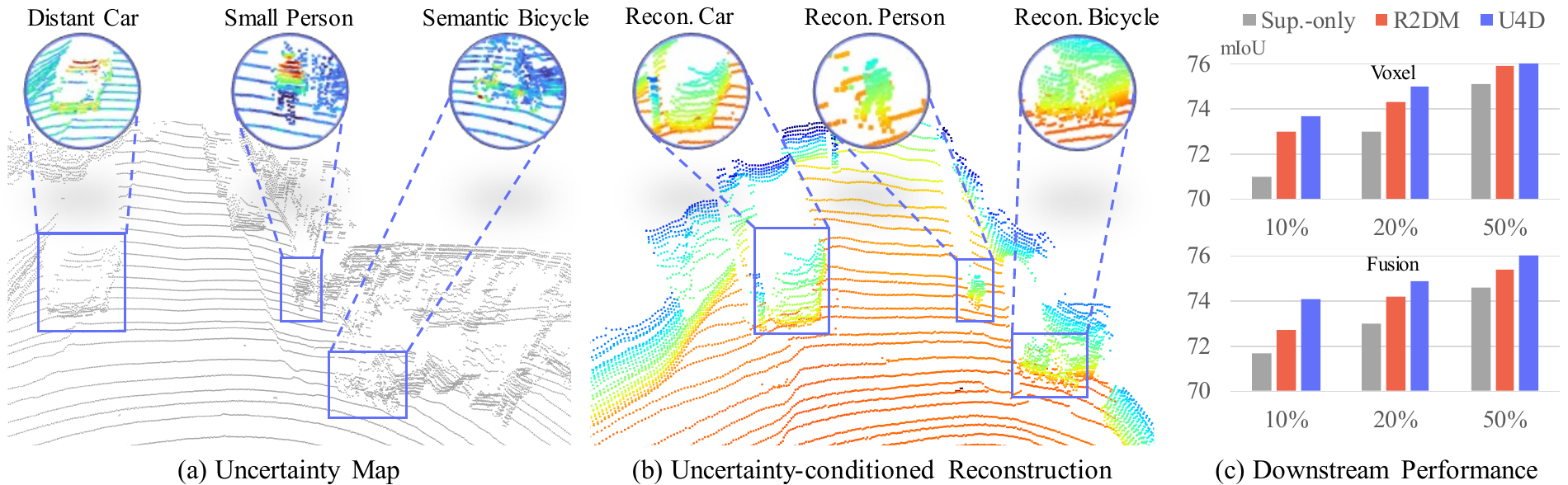}
    \vspace{-0.2cm}
    \caption{\textbf{Overview of U4D (Uncertainty-Aware 4D LiDAR Scene Synthesis).} \textbf{(a)} U4D estimates spatial uncertainty maps that highlight perceptually challenging regions such as distant objects, occluded structures, and semantically ambiguous areas. \textbf{(b)} Conditioned on these uncertainty regions, U4D performs scene generation in a ``hard-to-easy'' manner, progressively reconstructing the full scene with enhanced fidelity. \textbf{(c)} The resulting uncertainty-aware scenes improve downstream perception.}
    \label{fig:teaser}
    \vspace{0.2cm}
    \end{center}
}]

\begin{abstract}
Constructing faithful 4D worlds from LiDAR-acquired sequences is crucial for embodied AI, yet current generative frameworks apply uniform modeling capacity across all spatial regions. This ignores that perceptual difficulty varies dramatically within a single scan: distant surfaces, occluded boundaries, and small-scale objects carry far higher uncertainty than well-observed structures. We present \textbf{U4D}, a new framework that explicitly leverages spatial uncertainty to guide LiDAR scene generation in a ``hard-to-easy'' schedule. U4D derives per-point uncertainty maps via Shannon Entropy from a pretrained segmentor, then applies an unconditional diffusion stage to synthesize high-entropy areas with precise geometry, followed by a conditional completion stage that fills in the remaining regions using these structures as priors. A MoST (Mixture of Spatio-Temporal) block further maintains cross-frame coherence by dynamically balancing spatial detail and temporal continuity. Extensive experiments on nuScenes and SemanticKITTI demonstrate state-of-the-art scene fidelity, temporal consistency, and downstream performance.
\end{abstract}
\vspace{-0.2cm}
\section{Introduction}
\label{sec:intro}

Not all regions of a LiDAR scan are equally amenable to generation \cite{xu2026u4d}. A flat road surface is geometrically simple and semantically unambiguous, while a distant pedestrian partially occluded by a parked vehicle is sparse, boundary-heavy, and prone to semantic confusion \cite{kong2023lasermix,kong2025survey,kong2025lasermix++,worldlens}. Yet every existing LiDAR generation framework allocates identical modeling capacity to both, as if the entire scene were uniformly predictable \cite{zyrianov2022lidargen,wu2024text2lidar,agentic_world_modeling,liu2026la_la_lidar,liu2026omnilidar}. The consequence is systematic: faithful roads and buildings, but distorted poles, ghostly pedestrians, and flickering objects at range.

This observation motivates a fundamental shift in how we approach LiDAR world modeling. Rather than treating generation as a spatially uniform process, we argue that the model should first identify \emph{where} the scene is challenging, and then allocate its capacity accordingly. This ``hard-to-easy'' paradigm mirrors how humans resolve visual ambiguity: we fixate on uncertain regions first, anchoring our perception before integrating the surrounding context.

Building on this insight, we introduce \textbf{U4D}, a framework that brings uncertainty awareness to 4D LiDAR generation~\cite{xu2026u4d}. As shown in Fig.~\ref{fig:teaser}, U4D operates in three steps. First, it computes a spatial uncertainty map using Shannon Entropy~\cite{shannon1948entropy} over the predictions of a pretrained LiDAR segmentor, localizing areas of elevated semantic ambiguity such as occluded boundaries, thin structures, and distant surfaces. Second, it synthesizes scenes through two sequential diffusion stages: (1) an unconditional stage that reconstructs these high-entropy areas with precise geometric detail, and (2) a conditional completion stage that populates the remainder of the scene using the reconstructed uncertain structures as guiding priors. Third, to ensure temporal stability across frames, U4D incorporates a MoST (Mixture of Spatio-Temporal) block, a gating mechanism that dynamically balances spatial detail and temporal continuity within the diffusion backbone.

Extensive experiments on nuScenes~\cite{caesar2020nuscenes} and SemanticKITTI~\cite{behley2019semantickitti} validate that this uncertainty-guided strategy consistently outperforms uniform-generation baselines in scene-level fidelity ($6\%$--$11\%$ lower FRD/FPD), temporal coherence (lowest TTCE across all frame intervals), and downstream semantic segmentation. These results confirm that explicitly modeling \emph{where} a scene is challenging, before deciding \emph{how} to generate it, yields more reliable 4D LiDAR worlds.

\section{The U4D Framework}
\label{sec:method}

\begin{figure*}[t]
    \centering
    \includegraphics[width=\linewidth]{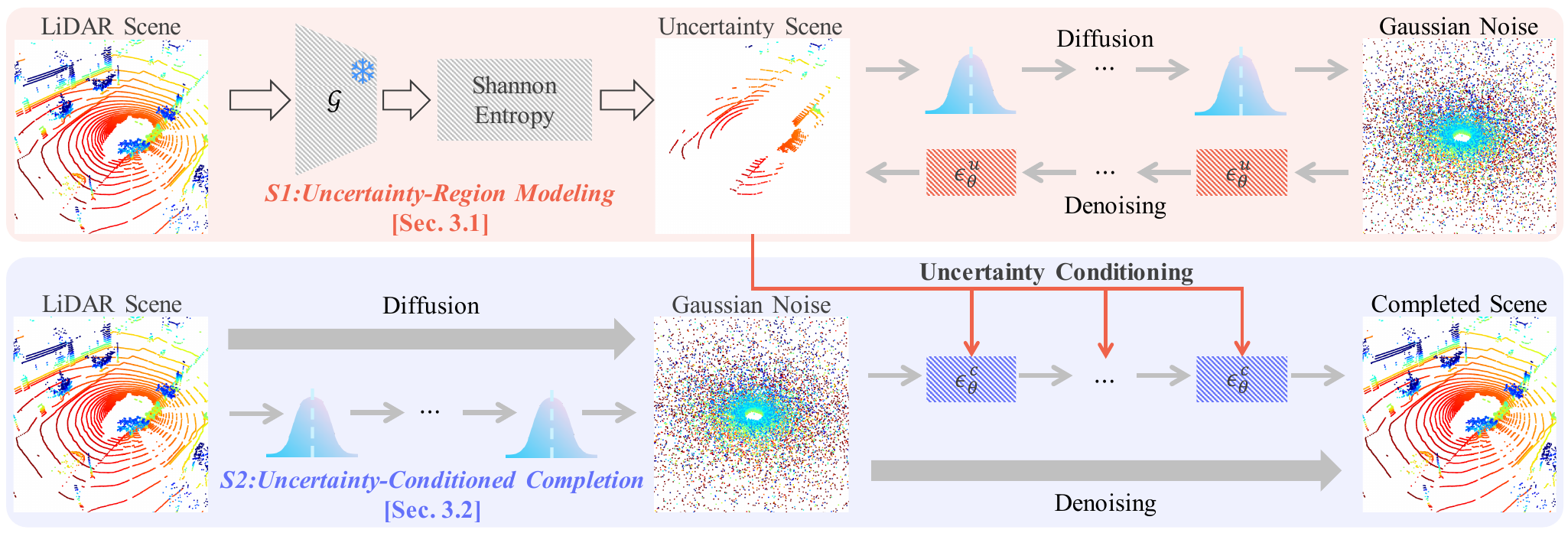}
    \vspace{-0.6cm}
    \caption{\textbf{U4D framework.} Stage~1: estimate uncertainty via Shannon Entropy, then reconstruct uncertain regions with unconditional diffusion. Stage~2: complete the full scene conditioned on the reconstructed structures.}
    \label{fig:framework}
    \vspace{-0.4cm}
\end{figure*}

U4D is grounded in the observation that LiDAR scenes exhibit spatially non-uniform difficulty, and that explicitly modeling this heterogeneity yields more faithful generation. The framework (Fig.~\ref{fig:framework}) comprises three core components: uncertainty measurement, two-stage hard-to-easy generation, and spatio-temporal fusion for cross-frame coherence.

\noindent\textbf{Spatial Uncertainty Estimation.}
Given a point cloud $\mathcal{P}$ and a pretrained LiDAR segmentation model $\mathcal{G}$, we extract per-point class probabilities $D_{c}(\mathbf{p})$ from the softmax-normalized logits and quantify semantic uncertainty via Shannon Entropy:
\begin{equation}
    H(\mathbf{p}) = -\sum\nolimits_{c=1}^{C} D_{c}(\mathbf{p}) \log D_{c}(\mathbf{p})~.
\end{equation}
Points with high entropy correspond to semantically ambiguous or geometrically unstable locations, typically near class boundaries, at long range, or within occluded areas. We select the top-$K$ highest-entropy points to construct a sparse point cloud of uncertain regions $\mathcal{P}^{u}$, which is then projected into a range-view representation $\mathbf{x}_{0}^{u} \in \mathbb{R}^{H \times W \times 2}$ encoding normalized depth and reflectance, along with a binary occupancy mask $\mathbf{m}^{u}$.

\noindent\textbf{Stage 1: Uncertainty-Region Generation.}
The first stage focuses exclusively on reconstructing geometry within the identified uncertain regions. An unconditional diffusion model $\epsilon_{\theta}^{u}$ is trained on range-view representations of uncertainty point clouds following the standard DDPM formulation~\cite{ho2020ddpm}. The training objective pairs a denoising loss with binary cross-entropy mask supervision:
\begin{equation}
    \mathcal{L}_{u} = \mathbb{E}_{t, \mathbf{x}_{0}^{u}, \epsilon^{u}} \left[\| \epsilon^{u} - \epsilon_{\theta}^{u}(\mathbf{x}_{t}^{u}, t) \|_{2}^{2} \right] + \lambda \mathcal{L}_{\mathrm{mask}}(\mathbf{m}^{u}, \mathbf{m}^{p})~,
\end{equation}
where $\mathcal{L}_{\mathrm{mask}}$ denotes binary cross-entropy and $\lambda$ controls its relative weight. At inference, iterative denoising from Gaussian noise yields a structured range image $\mathbf{\hat{x}}_{0}^{u}$ that captures fine-grained geometry in high-entropy areas while preserving valid sparsity patterns.

\noindent\textbf{Stage 2: Uncertainty-Conditioned Completion.}
While Stage~1 reconstructs structurally challenging areas, the scene remains incomplete. A conditional diffusion model $\epsilon_{\theta}^{c}$ learns to synthesize the full observation $\mathbf{x}_{0}$ given the uncertainty prior $\mathbf{x}_{0}^{u}$ by concatenating $\mathbf{x}_{t}$ with $\mathbf{x}_{0}^{u}$ along the channel dimension:
\begin{equation}
    \mathcal{L}_{c} = \mathbb{E}_{t, \mathbf{x}_{0}, \epsilon^{c}} \left[ \| \epsilon^{c} - \epsilon_{\theta}^{c}(\mathbf{x}_{t}, t, \mathbf{x}_{0}^{u}) \|_{2}^{2} \right]~.
\end{equation}
This conditioning forces the model to use the reconstructed uncertain areas as structural anchors, preserving occluded, distant, and small-scale elements in the completed scene. Both stages share a unified latent representation, enabling scene-level context to inform local uncertainty refinement.

\begin{figure}[t]
    \centering
    \includegraphics[width=\linewidth]{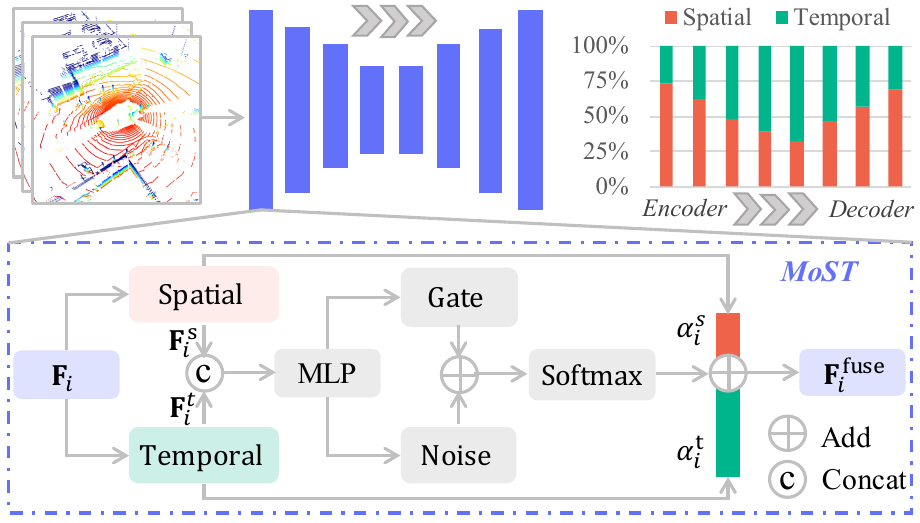}
    \vspace{-0.6cm}
    \caption{\textbf{MoST block.} Spatial and temporal branches are fused via adaptive gating. Spatial cues dominate near input/output; temporal dynamics dominate in intermediate layers.}
    \vspace{-0.2cm}
    \label{fig:most}
\end{figure}

\noindent\textbf{Mixture of Spatio-Temporal (MoST) Block.}
Temporal coherence across LiDAR frames is essential for realistic 4D generation yet challenging to maintain alongside spatial fidelity. The MoST block (Fig.~\ref{fig:most}) addresses this by decomposing intermediate features $\mathbf{F}_{i} \in \mathbb{R}^{C_{i} \times L \times H_{i} \times W_{i}}$ into a spatial branch $\mathbf{F}_{i}^{s}$ (intra-frame geometry via spatial convolution) and a temporal branch $\mathbf{F}_{i}^{t}$ (cross-frame alignment via temporal convolution). These are fused through a gating mechanism inspired by mixture-of-experts designs~\cite{xu2025limoe}:
\begin{equation}
    \mathbf{F}_{i}^{\mathrm{fuse}} = \alpha_{i}^{s} \odot \mathbf{F}_{i}^{s} + \alpha_{i}^{t} \odot \mathbf{F}_{i}^{t}~,
\end{equation}
where the gating weights $(\alpha_{i}^{s}, \alpha_{i}^{t})$ are computed from a shared MLP embedding with stochastic noise regularization during training. A coefficient-of-variation regularization term prevents the gating from degenerating to a single branch. Empirical analysis of the learned weights reveals an intuitive pattern: spatial cues dominate near input and output layers (governing geometric reconstruction), while temporal dynamics receive stronger weighting in intermediate layers (governing motion modeling).

\section{Experiments}
\label{sec:experiments}

We evaluate U4D on nuScenes~\cite{caesar2020nuscenes} and SemanticKITTI~\cite{behley2019semantickitti} across three dimensions: scene-level fidelity, temporal coherence, and downstream utility. Full implementation details and more results are provided in the main paper~\cite{xu2026u4d}.

\begin{figure*}[t]
    \centering
    \includegraphics[width=0.98\linewidth]{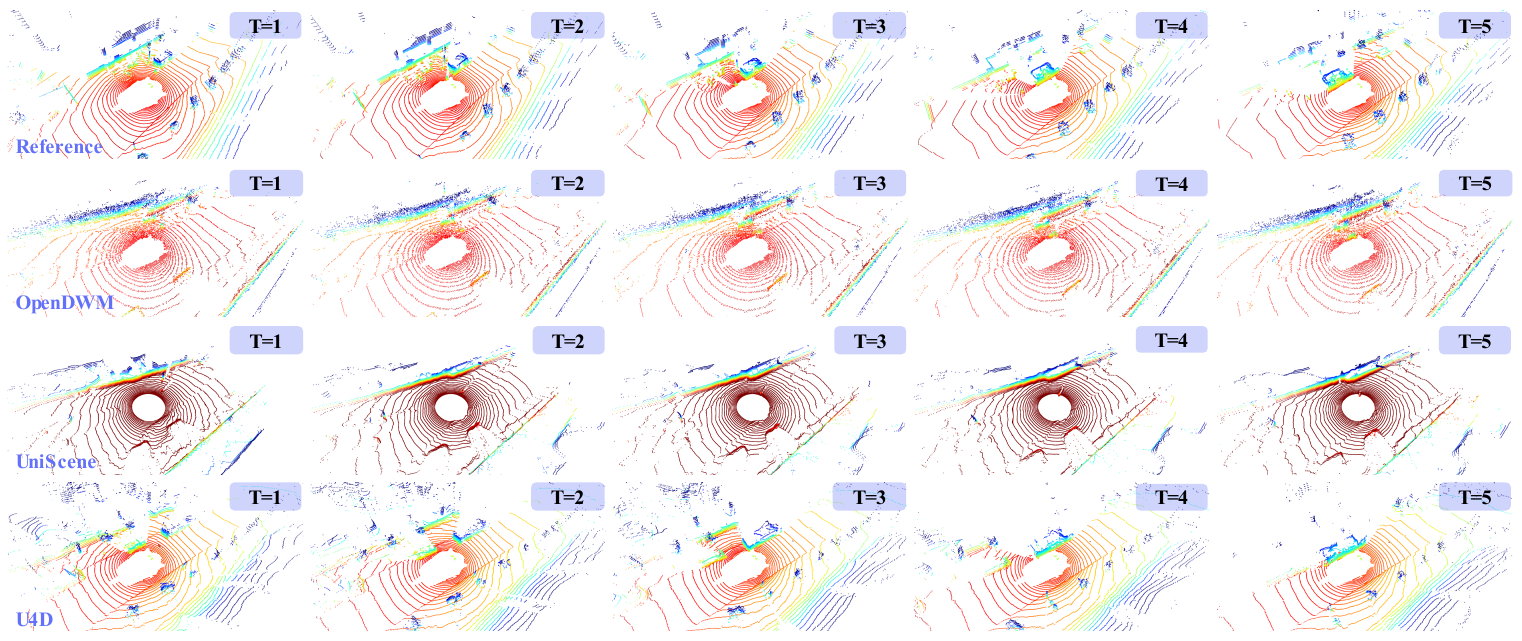}
    \vspace{-0.25cm}
    \caption{\textbf{Sequential point cloud generation} visualization results on the nuScenes~\cite{caesar2020nuscenes} dataset. U4D preserves geometric detail and temporal coherence across consecutive frames. Colors indicate point height.}
    \label{fig:vis_scene}
    \vspace{-0.3cm}
\end{figure*}

\begin{table}[t]
    \centering
    \caption{\textbf{LiDAR scene generation} on nuScenes~\cite{caesar2020nuscenes}. MMD in $10^{-4}$. \textbf{Bold}: best; \underline{underline}: second-best.}
    \vspace{-0.3cm}
    \label{tab:benchmark_nus}
    \resizebox{\linewidth}{!}{
    \begin{tabular}{r|r|cccc}
        \toprule
        \textbf{Method} & \textbf{Venue} & \textbf{FRD}~$\downarrow$ & \textbf{FPD}~$\downarrow$ & \textbf{JSD}~$\downarrow$ & \textbf{MMD}~$\downarrow$
        \\\midrule\midrule
        LiDARGen~\cite{zyrianov2022lidargen} & ECCV'22 & $549.18$ & $22.80$ & \underline{$0.04$} & $0.76$
        \\
        LiDM~\cite{ran2024lidm} & CVPR'24 & - & $30.77$ & $0.07$ & $3.86$
        \\
        R2DM~\cite{nakashima2024r2dm} & ICRA'24 & \underline{$253.80$} & \underline{$14.35$} & $\mathbf{0.03}$ & $\mathbf{0.48}$
        \\
        Text2LiDAR~\cite{wu2024text2lidar} & ECCV'24 & $953.18$ & $147.48$ & $0.09$ & $12.50$
        \\
        UniScene~\cite{li2025uniscene} & CVPR'25 & - & $976.47$ & $0.32$ & $13.61$
        \\
        OpenDWM~\cite{opendwm} & CVPR'25 & - & $714.19$ & $0.20$ & $5.61$
        \\\midrule
        \cellcolor{gen_blue!12}\textsl{U4D} & \cellcolor{gen_blue!12}\textbf{Ours} & \cellcolor{gen_blue!12}$\mathbf{223.96}$ & \cellcolor{gen_blue!12}$\mathbf{12.90}$ & \cellcolor{gen_blue!12}$\mathbf{0.03}$ & \cellcolor{gen_blue!12}\underline{$0.53$}
        \\\bottomrule
    \end{tabular}}
    \vspace{-0.2cm}
\end{table}

\begin{table}[t]
    \centering
    \caption{\textbf{Temporal consistency} on nuScenes~\cite{caesar2020nuscenes}. Numbers denote frame intervals. \textbf{Bold}: best; \underline{underline}: second-best.}
    \vspace{-0.3cm}
    \label{tab:benchmark_temporal}
    \resizebox{\linewidth}{!}{
    \begin{tabular}{r|r|cc|cccc}
        \toprule
        \multirow{2}{*}{\textbf{Method}} & \multirow{2}{*}{\textbf{Venue}} & \multicolumn{2}{c|}{\textbf{TTCE}~$\downarrow$} & \multicolumn{4}{c}{\textbf{CTC}~$\downarrow$}
        \\
        & & $\mathbf{3}$ & $\mathbf{4}$ & $\mathbf{1}$ & $\mathbf{2}$ & $\mathbf{3}$ & $\mathbf{4}$
        \\\midrule\midrule
        UniScene~\cite{li2025uniscene} & CVPR'25 & $2.74$ & $3.69$ & $\mathbf{0.90}$ & $\mathbf{1.84}$ & $3.64$ & $\mathbf{3.90}$
        \\
        OpenDWM~\cite{opendwm} & CVPR'25 & $2.68$ & $3.65$ & $1.02$ & $2.02$ & $3.37$ & $5.05$
        \\
        LiDARCrafter~\cite{liang2026lidarcrafter} & AAAI'26 & \underline{$2.65$} & \underline{$3.56$} & $1.12$ & $2.38$ & \underline{$3.02$} & $4.81$
        \\\midrule
        \cellcolor{gen_blue!12}\textsl{U4D} & \cellcolor{gen_blue!12}\textbf{Ours} & \cellcolor{gen_blue!12}$\mathbf{2.63}$ & \cellcolor{gen_blue!12}$\mathbf{3.51}$ & \cellcolor{gen_blue!12}\underline{$0.97$} & \cellcolor{gen_blue!12}\underline{$1.93$} & \cellcolor{gen_blue!12}$\mathbf{2.98}$ & \cellcolor{gen_blue!12}\underline{$4.41$}
        \\\bottomrule
    \end{tabular}}
    \vspace{-0.4cm}
\end{table}

\noindent\textbf{Scene-Level Fidelity.}
As shown in \cref{tab:benchmark_nus}, U4D achieves the strongest Fr\'{e}chet Range Distance ($223.96$) and Fr\'{e}chet Point Distance ($12.90$) on nuScenes, surpassing R2DM~\cite{nakashima2024r2dm} by roughly $6\%$--$11\%$. It also matches or exceeds all baselines on BEV-based metrics (JSD, MMD), confirming robust spatial consistency across viewpoints. On SemanticKITTI, comparable gains are observed, with U4D achieving $245.73$ FRD and $10.92$ FPD. These improvements stem from the uncertainty-guided strategy: by first reconstructing challenging regions as structural anchors, the subsequent completion stage benefits from stronger geometric priors, particularly for distant and occluded surfaces that uniform approaches reconstruct poorly.

\noindent\textbf{Temporal Coherence.}
Generating temporally stable LiDAR sequences is essential for simulation, yet inconsistent frame-to-frame evolution remains a prevalent failure mode. As reported in \cref{tab:benchmark_temporal}, U4D achieves the lowest TTCE across all measured frame intervals ($2.63$ at interval 3, $3.51$ at interval 4), indicating that the generated sequences exhibit smooth and plausible ego-motion patterns. Chamfer Temporal Consistency (CTC) scores are also competitive, reflecting stable point cloud geometry between adjacent frames. The MoST block is the primary contributor: its adaptive spatial-temporal gating ensures that geometric detail and motion continuity are jointly maintained rather than traded off against each other.

\noindent\textbf{Downstream Perception.}
The practical value of generated LiDAR data depends on whether it improves real-world perception models. Following LaserMix~\cite{kong2023lasermix}, we employ U4D-generated scenes as unlabeled augmentation data. Across both voxel~\cite{choy2019minkunet} and fusion~\cite{tang2020spvcnn} backbones, incorporating U4D data consistently improves over baselines trained with semi-supervised methods alone or with R2DM-generated augmentation. Also, evaluation using the Expected Calibration Error (ECE) metric~\cite{kong2025calib3d} confirms that uncertainty-aware generation enhances model calibration, yielding more reliable prediction confidence alongside higher accuracy.

\noindent\textbf{Ablation Highlights.}
(1) \emph{Uncertainty selection}: entropy-based region selection outperforms both no-conditioning and random sampling by a clear margin, and also surpasses confidence-based alternatives, confirming that entropy effectively localizes varied and informative regions of difficulty. (2) \emph{MoST design}: adaptive gating fusion outperforms cascaded, additive, and concatenation-based spatial-temporal fusion strategies, validating the mixture-of-experts formulation. (3) \emph{Activation analysis}: the learned gating weights indicate that spatial cues dominate near network boundaries while temporal dynamics peak in intermediate layers, consistent with the intuition that reconstruction and motion modeling operate at distinct abstraction levels.

\section{Conclusion}
\label{sec:conclusion}

We presented U4D, an uncertainty-aware generative framework that reframes LiDAR scene synthesis as a spatially adaptive process. By deriving per-point semantic uncertainty via Shannon Entropy and generating scenes in a hard-to-easy order, U4D concentrates modeling capacity on the regions that matter most: occluded boundaries, distant surfaces, and semantically ambiguous structures. The Mixture of Spatio-Temporal block further ensures that this spatial precision extends coherently across time. Experiments on nuScenes and SemanticKITTI confirm consistent improvements in scene-level fidelity, cross-frame stability, and downstream segmentation utility, demonstrating that explicit awareness of scene difficulty serves as a practical and effective prior for reliable 4D LiDAR world modeling.

{
    \small
    \bibliographystyle{abbrv}
    \bibliography{main}
}

\end{document}